\documentclass[conference]{IEEEtran}
\usepackage{cite}

\ifCLASSINFOpdf
  \usepackage[pdftex]{graphicx}
\else
   \usepackage[dvips]{graphicx}
\fi
\usepackage{amsmath}
\usepackage[ruled,linesnumbered]{algorithm2e}

\SetKwProg{Fn}{Function}{}{}
\SetKwRepeat{Struct}{type}{}%
\SetKw{This}{this}
\SetKw{False}{false}
\SetKw{True}{true}
\SetKw{Null}{null}
\SetKw{Type}{type}
\SetKwComment{Comment}{$\triangleright$\ }{}
\IncMargin{1em}
\everypar={\;}
 \usepackage{listings}
 \usepackage{xcolor}
\definecolor{britishracinggreen}{rgb}{0.0, 0.26, 0.15}
\ifCLASSOPTIONcompsoc
  \usepackage[caption=false,font=normalsize,labelfont=sf,textfont=sf]{subfig}
\else
  \usepackage[caption=false,font=footnotesize]{subfig}
\fi

\usepackage{stfloats}

\usepackage{multicol}

\begin{document}
%
\title{Structured Parallel Programming for Monte Carlo Tree Search}

\author{\IEEEauthorblockN{S. Ali Mirsoleimani\IEEEauthorrefmark{1}\IEEEauthorrefmark{2},
Aske Plaat\IEEEauthorrefmark{1},
Jaap van den Herik\IEEEauthorrefmark{1} and
Jos Vermaseren\IEEEauthorrefmark{2}}
\IEEEauthorblockA{\IEEEauthorrefmark{1}Leiden Centre of Data Science, Leiden University\\
 Niels Bohrweg 1, 2333 CA Leiden, The Netherlands}
\IEEEauthorblockA{\IEEEauthorrefmark{2}Nikhef Theory Group, Nikhef\\
 Science Park 105, 1098 XG Amsterdam, The Netherlands}}


%


\maketitle

\begin{abstract}
 
 
In this paper, we present a new algorithm for parallel Monte Carlo tree search (MCTS). It is based on the pipeline pattern and allows flexible management of the control flow of the operations in parallel MCTS. The pipeline pattern provides for the first structured parallel programming approach to MCTS. Moreover, we propose a new lock-free tree data structure for parallel MCTS which removes synchronization overhead. The Pipeline Pattern for Parallel MCTS algorithm (called 3PMCTS), scales very well to higher numbers of cores when compared to the existing methods.
\end{abstract}

%
\IEEEpeerreviewmaketitle

\section{Introduction}
In recent years there has been much interest in the Monte Carlo tree search (MCTS) algorithm, at that time a new, adaptive, randomized optimization algorithm \cite{Coulom2006,Kocsis2006}. In fields as diverse as Artificial Intelligence, Operations Research, and High Energy Physics, research has established that MCTS can find valuable approximate answers without domain-dependent heuristics \cite{Kuipers2013}. The strength of the MCTS algorithm is that it provides answers with a random amount of error for any fixed computational budget \cite{goodfellow2016deep}. The amount of error can typically be reduced by expanding the computational budget for more running time. Much effort has been put into the development of parallel algorithms for MCTS to reduce the running time. The efforts have as their target a broad spectrum of parallel systems; ranging from small shared-memory multicore machines (CPU) to large distributed-memory clusters. The emergence of the Intel Xeon Phi (Phi) co-processor with a large number (over 60) of simple cores has extended this spectrum with shared-memory manycore processors. Indeed, there is a full array of different parallel MCTS algorithms \cite{Chaslot2008,Yoshizoe2011a,fern2011ensemble,Schaefers2014,Mirsoleimani2015a,Mirsoleimani2015}. However, there is still no \textit{structured parallel programming} approach, based on computation patterns, for MCTS. In this paper, we propose a new algorithm based on the \textit{Pipeline Pattern} for Parallel MCTS, called 3PMCTS.  

The standard MCTS algorithm has four operations inside its main loop (Figure \ref{fig:mcts-flow}). In this loop, the computation associated with each iteration is assumed to be independent. Existing parallel methods use \textit{iteration-level} (IL) parallelism. They assign each iteration to a separate processing element (thread) for execution on separate processors \cite{Chaslot2008,Schaefers2014,Mirsoleimani2015a}. All three publications are facing a bottleneck in their implementation since they can not partition the iteration into constituent parts (operations) for parallel execution. Close analysis has learned us that the loop can actually be decomposed into separate operations for parallelization. Therefore, in our new design we introduce \textit{operation-level} (OL) parallelism. The main idea is that the 3PMCTS algorithm assigns each operation to a separate processing element for execution by separate processors. This leads to flexibility in managing the control flow of operations in the MCTS algorithm.

Our approach of applying structured parallel programming focuses the attention on (1) a design issue and (2) an implementation issue. For the design issue, we describe how the pipeline pattern is used as a building block in the design of 3PMCTS. It consists of a precise arrangement of tasks and data dependencies in MCTS. 
For the implementation, it is important to measure the performance of 3PMCTS on real machines. We present the implementation of 3PMCT by using Threading Building Blocks (TBB) \cite{tbb2015} and we measure its performance on CPU and Phi.\footnote{We also discuss two elements related to the implementation of 3PMCTS, including the concept of \textit{token} and a new lock-free tree data structure. A lock-free tree data structure plays a critical role in any parallel implementation for MCTS to be scalable.} This paper has three contributions:
\begin{enumerate}
\item A new structured algorithm based on the pipeline pattern is introduced for parallel MCTS. 
\item A new lock-free tree data structure for parallel MCTS is presented.
\item A new TBB implementation based on the concept of \textit{token} is proposed. The experimental results show that our implementation scales well.
\end{enumerate}

The remainder of the paper is organized as follows. In section \ref{sec:back} the required background information is briefly described. Section \ref{sec:sematic_3pmcts} provides necessary definitions and explanations for the design issue of structured parallel programming for 3PMCTS. Section \ref{sec:implement} gives the explanations for the implementation issue of the 3PMCTS algorithm. Section \ref{sec:implement_lock_free_tree} provides the proposed lock-free tree data structure, Section \ref{sec:setup} the experimental setup of the 3PMCTS, and Section \ref{sec:results} the experimental results for 3PMCTS. Section \ref{sec:related} discusses related work. Finally, in Section \ref{sec:conclusion} we conclude the paper.

\section{Background}
Below we discuss in \ref{sec:back_mcts} MCTS, in \ref{sec:treepar} tree parallelization, and in \ref{sec:tbb} TBB.
\label{sec:back}
\subsection{MCTS}
\label{sec:back_mcts}
\begin{figure}
\centering
\includegraphics[scale=0.5]{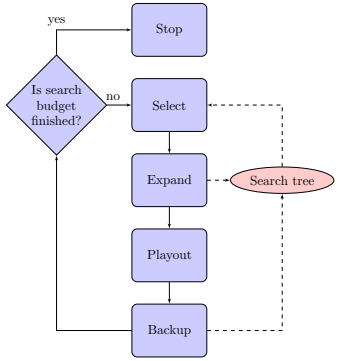}
\caption{Flowchart of the sequential MCTS.}
\label{fig:mcts-flow}
\end{figure}

Figure \ref{fig:mcts-flow} shows a flowchart of MCTS \cite{Coulom2006}. The MCTS algorithm iteratively repeats four steps to construct a search tree until a predefined computational budget (i.e., time or iteration constraint) is reached. 
\begin{enumerate}
\item \textsc{Select}: Starting at the root node, a path of nodes inside the search tree is selected until a non-terminal node with unvisited children is reached. Each of the nodes is selected based on a \textit{selection policy}. Among the proposed selection policies, the Upper Confidence Bounds for Trees (UCT) is one of the most commonly used policies~\cite{Browne2012,Kocsis2006}. A child node $j$ is selected to maximize: 
\begin{equation}
UCT=\overline{X}_{j}+C_{p}\sqrt{\frac{\ln(n)}{n_{j}}}
\end{equation}
where $\overline{X}_{j}=\frac{w_{j}}{n_{j}} $ is the average reward from child $j$, $w_{j}$ is the reward value
of child $j$, $n_{j}$ is the number of times child $j$ has
been visited, $n$ is the number of times the current node has been
visited, and $C_{p}\geq0$ is a constant. The first term in the UCT
equation is for \textit{exploitation} of known parts of the tree and the second term is for
\textit{exploration} of unknown parts~\cite{Browne2012}. The level of exploration of the UCT
algorithm can be adjusted by tuning the $C_{p}$ constant.

\item \textsc{Expand}: One of the children of the selected non-terminal node is generated and added to the selected path.
\item \textsc{Playout}: From the given state of the newly added node, a sequence of randomly simulated actions is performed until a terminal state in the state space is reached. The terminal state is evaluated to produce a reward value $\Delta$.
\item \textsc{Backup}: In the selected path, each node's visit count $n$ is incremented by 1 and its reward value $w$ updated according to $\Delta$ \cite{Browne2012}.
\end{enumerate}
As soon as the computational budget is exhausted, the best child of the root node is returned.

\subsection{Tree Parallelization}
\label{sec:treepar}
\begin{figure}
\centering
\includegraphics[scale=1]{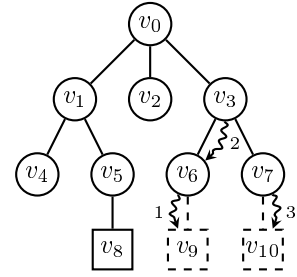}
\caption{Tree parallelization without locks. The curly arrows represent threads. The rectangles are terminal leaf nodes.}
\label{fig:PMCTS-1}
\end{figure}

In tree parallelization, one search tree is shared among several threads that are performing simultaneous searches~\cite{Chaslot2008}. The main challenge in this method is the prevention of data corruption. A lock-based method uses fine-grained locks to protect shared data. However, this approach suffers from synchronization overhead due to thread contentions and does not scale well \cite{Chaslot2008}. A lock-free implementation of addresses the problem and scales better~\cite{Enzenberger2010a}. However, the method in~\cite{Enzenberger2010a} does not guarantee the computational consistency of the multi-threaded program with the single-threaded program. Figure \ref{fig:PMCTS-1} shows the tree parallelization without locks.

\subsection{TBB}
\label{sec:tbb}
 TBB is a C++ template library
developed by Intel for writing software programs that take advantage
of a multicore processor~\cite{tbb-reinders2007-book}. 
The TBB implementation of pipelines uses a technique that enables greedy scheduling, but the greed must be constrained to bound memory consumption. The user specifies the constraint as a maximum number of items allowed to flow simultaneously through the pipeline~\cite{tbb-reinders2007-book}. 

\section{Design of 3PMCTS}
\label{sec:sematic_3pmcts}
In this section, we describe our structured parallel programming approach for MCTS. In section \ref{sec:decomp_task} we explain how to decompose MCTS into tasks. In section \ref{sec:data_depen} we investigate what types of data dependencies exist among these tasks. Section \ref{sec:pipeline_for_mcts} describes how the pipeline pattern is applied in MCTS. Finally, section \ref{sec:semantic_parallelism} provides our design for the 3PMCTS algorithm.

\subsection{Decomposition into Tasks}
\label{sec:decomp_task}
The first step towards designing our 3PMCTS algorithm is to find concurrent tasks in MCTS. There are two levels of \textbf{task decomposition} in MCTS.
\begin{enumerate}

\item \textit{Iteration-level (IL) tasks}: In MCTS the computation associated with each \textsc{Select}-\textsc{Expand}-\textsc{Playout}-\textsc{Backup}-iteration is independent. Therefore, {\it these are candidates} to guide a task decomposition by mapping each iteration onto a task. 
\item \textit{Operation-level (OL) tasks}: The task decomposition for MCTS occurs inside each iteration. Each of the four MCTS steps can be treated as a separate task. 

\end{enumerate}
\subsection{Data Dependencies}
\label{sec:data_depen}
In 3PMCTS, a search tree is shared among multiple parallel tasks. Therefore, there are two levels of \textbf{data dependency}.

\begin{enumerate}
\item \textit{Iteration-level (IL) dependencies}: Strictly speaking, in MCTS, iteration $j$ has a \textit{soft dependency} to its predecessor iteration $j-1$. Obviously, to select an optimal path, it requires updates on the search tree from the previous iteration.\footnote{i.e., a violation of IL dependency does not impact the correctness of the algorithm.} A parallel MCTS can ignore IL dependencies and simply suffers from the \textit{search overhead}.\footnote{Occurs when a parallel implementation in a search algorithm searches more nodes of the search space than the sequential version; for example, since the information to guide the search is not yet available.} 
\item \textit{Operation-level (OL) dependencies}: Each of the four operations in MCTS has a \textit{hard dependency} to its predecessor.\footnote{i.e., a violation of OL dependency yields an incorrect algorithm.} For example, the expansion of a path cannot start until the selection operation has been completed.
\end{enumerate}

\subsection{Pipeline Pattern}
\label{sec:pipeline_for_mcts}
In this section, we focus on the pipeline pattern in MCTS using OL tasks. The pipeline pattern is the most straightforward way to enforce the required ordering among the OL tasks. Below we explain two possible types of pipelines for MCTS.

\begin{figure}[!t]
\begin{tabular}{cc}
\subfloat[]{\includegraphics[scale=0.5]{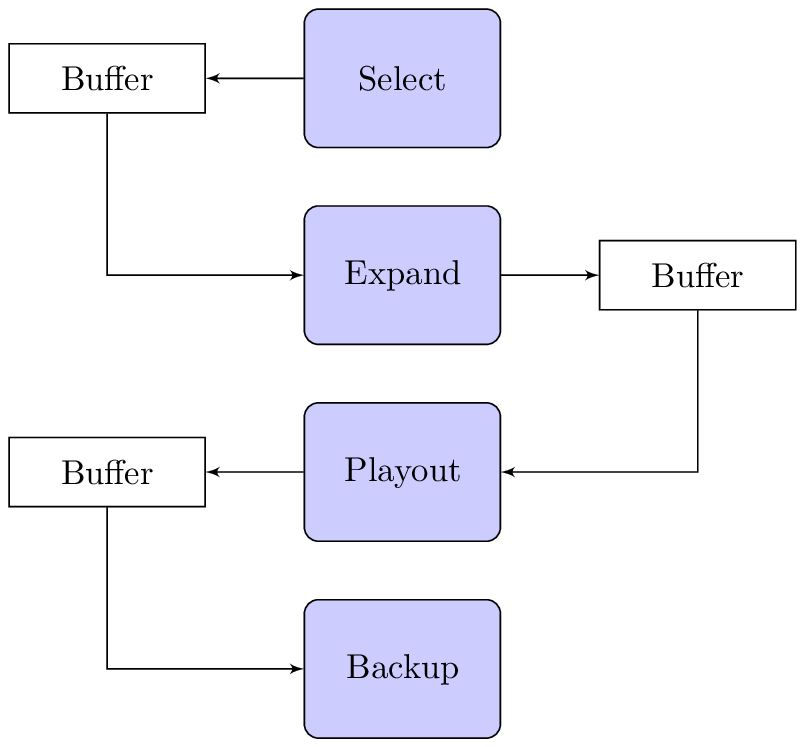}%
\label{fig:mcts-pipe}}
&
\subfloat[]{\includegraphics[scale=0.5]{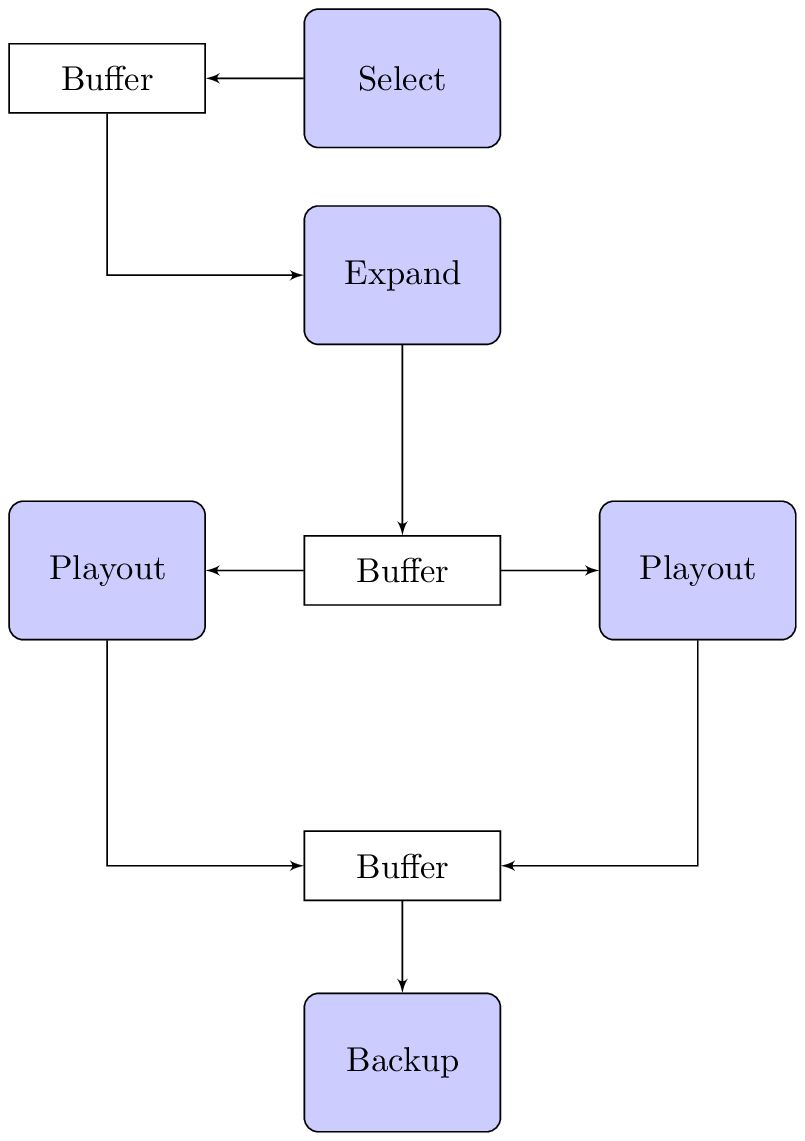}%
\label{fig:mcts-pipe-chart-multi}}
\end{tabular}
\caption{(\ref{fig:mcts-pipe}) Flowchart of a linear pipeline for MCTS. (\ref{fig:mcts-pipe-chart-multi}) Flowchart of a nonlinear pipeline for MCTS.}
\end{figure}

\begin{enumerate}
\item \textit{Linear pipeline}: Figure \ref{fig:mcts-pipe} shows a linear MCTS pipeline with the selected paths inside the search tree; from one stage to the next stage buffers are given a path as operations are completed.
\item \textit{Non-linear pipeline}: Figure \ref{fig:mcts-pipe-chart-multi} shows a non-linear MCTS pipeline with two parallel \textsc{Playout} stages. Both of the \textsc{Playout} stages take paths produced by the \textsc{Expand} stage of the pipeline.
\end{enumerate}

 
\subsection{Parallelism of a Pipeline}
\label{sec:semantic_parallelism}

The existing parallel methods such as tree parallelization use IL tasks. There are only IL dependencies when performing IL parallelism. The potential concurrency is exploited by assigning each of the IL tasks to a separate processing element and having them work on separate processors. So far IL parallelism is investigated quite extensively \cite{Chaslot2008,Enzenberger2010a,Schaefers2014,Mirsoleimani2015,Mirsoleimani2015a}. 

\begin{figure}[!t]
\centering
\includegraphics[scale=0.4]{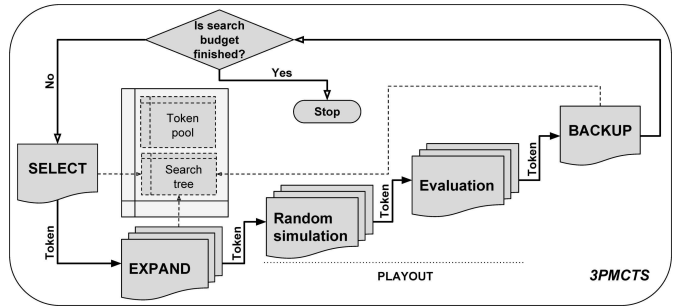}
\caption{The 3PMCTS algorithm with a five-stage non-linear pipeline.}
\label{fig:3pmcts}
\end{figure}

In contrast to the existing methods, our 3PMCTS algorithm uses OL tasks which have both IL and OL dependencies. The OL tasks are assigned to the stages of a pipeline. The pipeline pattern can satisfy the OL dependencies among the OL tasks. The potential concurrency is also exploited by assigning each stage of the pipeline to a separate processing element for execution on separate processors. If the pipeline is \textit{linear} then the scalability is limited to the number of stages.\footnote{When the operations performed by the various stages are all about equally computationally intensive.} However in MCTS, the operations are not equally computationally intensive, e.g., generally the \textsc{Playout} operation (random simulations plus evaluation of a terminal state) is more computationally expensive than other operations. Therefore, our 3PMCTS algorithm uses a \textit{non-linear} pipeline with parallel stages. Introducing parallel stages makes 3PMCTS more scalable. The 3PMCTS algorithm, depicted in Figure \ref{fig:3pmcts}, has three parallel stages (i.e., \textsc{EXPAND}, \textit{Random Simulation}, and \textit{Evaluation}). It will be usable in almost any sufficiently powerful parallel programming model  (e.g., TBB \cite{tbb-reinders2007-book} or Cilk \cite{cilk1998}).


\section{Implementation of 3PMCTS}
\label{sec:implement}
In this section, we introduce the implementation of our 3PMCTS algorithm. In section \ref{sec:implement_token} we present the concept of \textit{token} (when used as type name, we write \textit{Token}). Section \ref{sec:implement_3pmcts} describes the implementation of 3PMCTS using TBB.

\subsection{Token}
\label{sec:implement_token}
A token represents a path inside the search tree during the search. Algorithm \ref{algo:def_token} presents definition for the type \textit{Token}. It has four fields. (1) $id$ represents a unique identifier for a token, (2) \textit{v} represents the current node in the tree, (3) \textit{s} represents the search state of the current node, and (4) $\Delta$ represents the reward value of the state. In our implementation for 3PMCTS, each stage (task) performs its operation on a token. We can also specify the number of in-flight tokens.

\begin{algorithm}[!t]
\LinesNumbered
\Struct{Token}{

\Type $id$ : int\;
\Type $v$ : Node*\;
\Type $s$ : State*\;
\Type $\Delta$ : int\;

}
\caption{Type definition for token.}
\label{algo:def_token}
\end{algorithm}

\begin{algorithm}[!t]
\Fn{\textsc{UCTSearch}($s_{0}$)}{
$v_{0}$ = create root node with state $s_{0}$\;
$t_{0}.s$ = $s_{0}$\;
$t_{0}.v$ = $v_{0}$\;
\While{within search budget}{
$t_{l}$ = \textsc{Select}($t_{0}$)\;
$t_{l}$ = \textsc{Expand}($t_{l}$)\;
$t_{l}$ = \textsc{Playout}($t_{l}$)\;
\textsc{Backup}($t_{l}$)\;
}
}
\caption{Serial implementation of MCTS, with stages \textsc{Select}, \textsc{Expand}, \textsc{Playout}, and \textsc{Backup}.}
\label{algo:mcts-serial}
\end{algorithm}

\begin{algorithm}[!t]
\Fn{\textsc{Select}(Token $t$) : $<$Token$>$}{
\While{$t.v\rightarrow$IsFullyExpanded()}{
$t.v$ $:=$ $\underset{v^{'}\in children of v}{argmax}v^{'}$.UCT($C_{p}$)\;
$t.s\rightarrow$SetMove($t.v\rightarrow move$)\;
}
\Return $t$\;
}
\BlankLine
\Fn{\textsc{Expand}(Token $t$) : $<$Token$>$}{
\If{!($t.s\rightarrow$IsTerminal())}{
$moves :=$ $t.s\rightarrow$UntriedMoves()\;
shuffle $moves$ uniformly at random\;

$t.v\rightarrow$Init($moves$)\;
$v^{'} :=$ $t.v\rightarrow$AddChild()\;

\If{$t.v$ $\neq$ $v^{'}$}{
$t.v :=$$v^{'}$\;
$t.s\rightarrow$SetMove($v^{'}\rightarrow move$)\;
}
}
\Return t;
}
\BlankLine
\BlankLine
\Fn{RandomSimulation(Token $t$)}{
$moves :=$ $t.s\rightarrow$UntriedMoves()\;
shuffle $moves$ uniformly at random\;
\While{!($t.s\rightarrow$IsTerminal())}{
choose new $move\in moves$\;
$t.s\rightarrow$SetMove($move$)\;
}
\Return $t$
}
\BlankLine
\Fn{Evaluation(Token $t$)}{
$t.\Delta :=$ $t.s \rightarrow$ Evalute()\;
\Return $t$
}
\BlankLine
\Fn{\textsc{Backup}(Token $t$) : void}{
\While{$t.v$ $\neq$ \Null}{
$t.v \rightarrow Update$($t.\Delta$)\;
$t.v :=$ $t.v \rightarrow parent$\;
}
}
\caption{The functions of the MCTS algorithm.}
\label{algo:mcts-functions}
\end{algorithm}

\subsection{TBB Implementation}
\label{sec:implement_3pmcts}
The pseudocode of MCTS is shown in Algorithm \ref{algo:mcts-serial}. A data structure of type \textit{State} describes the search state of the current node in the tree and a data structure of type \textit{Node} shows the current node being searched inside the tree. The functions of the MCTS algorithm are defined in Algorithm \ref{algo:mcts-functions}. Each function constitutes a stage of the non-linear pipeline in 3PMCTS. There are two approaches for parallel implementation of a non-linear pipeline \cite{McCool2012}:
\begin{itemize}
\item \textit{Bind-to-stage}: A processing element (e.g., thread) is bound to a stage and processes tokens as they arrive. If the stage is parallel, it may have multiple processing elements bound to it. 
\item \textit{Bind-to-item}: A processing element is bound to a token and carries the token through the pipeline. When the processing element completes the last stage, it goes to the first stage to select another token. 
\end{itemize}
 
Our implementation for 3PMCTS algorithm is based on a bind-to-item approach. Figure \ref{fig:3pmcts} depicts a five-stage pipeline for 3PMCTS that can be implemented using TBB \textit{tbb::parallel\_pipeline} template \cite{tbb-reinders2007-book}. The five stages run the functions \textsc{Select}, \textsc{Expand}, \textit{RandomSimulation}, \textit{Evaluation}, and \textsc{Backup}, in that order. The first (\textsc{Select}) and last stage (\textsc{Backup}) are serial in-order; They process one token at a time.  The three middle stages (\textsc{Expand}, \textit{RandomSimulation},  and \textit{Evaluation}) are parallel and do the most time-consuming part of the search. The \textit{Evaluation} and \textit{RandomSimulation} functions are extracted out of the \textsc{Playout} function to achieve more parallelism. The serial version uses a single token. The 3PMCTS algorithm aims to search multiple paths in parallel. Therefore, it needs more than one in-flight \textit{token}.   
Figure \ref{lst:mcts-pipeline-sps} shows the key parts of the TBB code with the syntactic details for the 3PMCTS algorithm. 

\begin{figure*}[!t]
\begin{lstlisting}[xleftmargin=20pt,xrightmargin=0pt]
void 3PMCTS(tokenlimit){
...
/* The routine tbb::parallel_pipeline takes two parameters. 
(1) A token limit. It is an upper bound on the number of tokens that are processed simultaneously. 
(2) A pipeline. Each stage is created by function tbb::make_filter. The template arguments to 
make_filter indicate the type of input and output items for the filter. The first ordinary argument
specifies whether the stage is parallel or not and the second ordinary argument specifies a function
that maps the input item to the output item. 
*/
    tbb::parallel_pipeline(tokenlimit,
       /* The SELECT stage is serial and mapping a special object of type tbb::flow_control, used
       to signal the end of the search, to an output token. */ 
       tbb::make_filter<void, Token*>(tbb::filter::serial_in_order,[&](tbb::flow_control & fc)->Token* {
            /* A circular buffer is used to minimize the overhead of allocating and freeing tokens
            passed between pipeline stages (it reduces the communication overhead). */
            Token* t = tokenpool[index];
            index = (index+1) % tokenlimit;
            if (within the search budget) {
                /* Invocation of the method stop() tells the tbb::parallel_pipeline that no more 
                paths will be selected and that the value returned from the function should be 
                ignored. */
                fc.stop();
                return NULL;
            } else {
                t = SELECT(t);
                return t
            }
        }
        ) &
        // The EXPAND stage is parallel and mapping an input token to an output token.
        tbb::make_filter<Token*, Token*>(tbb::filter::parallel,[&](Token * t){ 
             return EXPAND(t);
        }) &
        // The RandomSimulation stage is parallel and mapping an input token to an output token. 
        tbb::make_filter<Token*, Token*>(tbb::filter::parallel,[&](Token * t){  
             return RandomSimulation(t);
        }) &
        // The Evaluation stage is parallel and mapping an input token to an output token. 
        tbb::make_filter<Token*, Token*>(tbb::filter::parallel,[&](Token * t){  
             retun Evaluation(t);
        }) &
        /* The  BACKUP stage has an output type of void since it is only consuming tokens, 
        not mapping them. */ 
        tbb::make_filter<Token*, void>(tbb::filter::serial_in_order,[&](Token * t){  
             return BACKUP(t);
        })
    );    
...      
}
\end{lstlisting}

\caption{An implementation of the 3PMCTS algorithm in TBB.} 
\label{lst:mcts-pipeline-sps}
\end{figure*}

\section{Lock-free Search Tree}
\label{sec:implement_lock_free_tree}
\begin{figure}[!t]
\centering
\scalebox{0.85}{
\begin{tabular}{ccccc}
\subfloat[]{\includegraphics[width=1.7in]{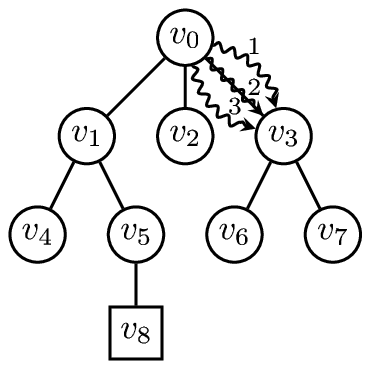}%
\label{fig:dr_ss_1}}
&
\subfloat[]{\includegraphics[width=1.7in]{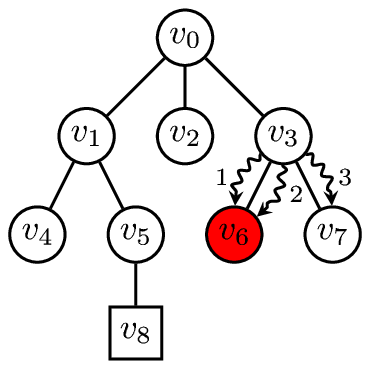}%
\label{fig:dr_se_1}}
\\
\subfloat[]{\includegraphics[width=1.7in]{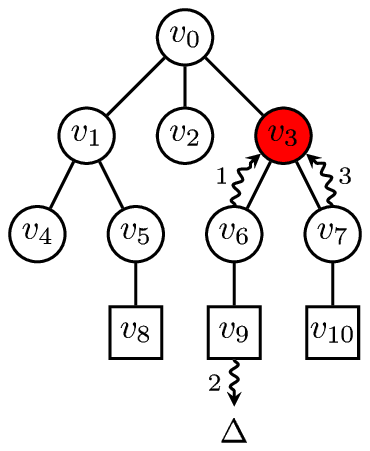}%
\label{fig:dr_sb_1}}
&
\subfloat[]{\includegraphics[width=1.7in]{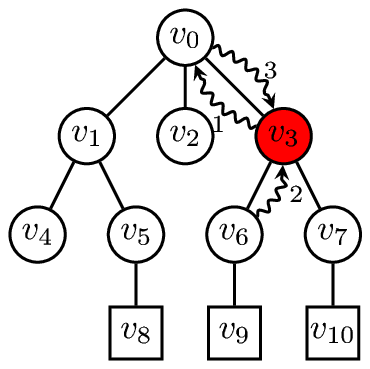}%
\label{fig:dr_sbs_1}}
\end{tabular}
}
\caption{(\ref{fig:dr_ss_1}) The initial search tree. The internal and non-terminal leaf nodes are circles.
The terminal leaf nodes are squares. (\ref{fig:dr_se_1}) Thread 1 and 2 are expanding node $v_{6}$. (\ref{fig:dr_sb_1}) Thread 1 and 2 are updating node $v_{3}$. (\ref{fig:dr_sbs_1}) Thread 1 is selecting node $v_{3}$ while thread 2 is updating this node.}
\label{fig:dr}
\end{figure}

In this section, we provide our new lock-free tree search. A potential bottleneck in a parallel implementation is the \textit{race condition}. A race condition occurs when concurrent threads perform operations on the same memory location without proper synchronization, and one of the memory operations is a write \cite{McCool2012}. Consider the example search tree in Figure \ref{fig:dr_ss_1}. Three parallel threads attempt to perform MCTS operations on the shared search tree. There are three race condition scenarios.

\begin{enumerate}
\item Shared Expansion (SE): Figure \ref{fig:dr_se_1} shows two threads ($1$ and $2$) concurrently performing \textsc{Expand}($v_{6}$). In this SE scenario, synchronization is required. Obviously, a race condition exists if two parallel threads intend to initialize the list of children in a node simultaneously. In such an SE race, the list of children for a selected node should be created only once. Enzenberger et al. assign to each thread an own memory array for creating a list of new children \cite{Enzenberger2010b}. Only after the children are fully created and initialized, they are linked to the parent node. Of course, this causes memory overhead. What usually happens is the following. If several threads expand the same node, only the children created by the last thread will be used in future simulations. It can also happen that some of the children that are lost already received some updates; these updates will also be lost.  
\item Shared Backup(SB): Figure \ref{fig:dr_sb_1} shows two threads ($1$ and $3$) concurrently performing \textsc{Backup}($v_{3}$). In the SB scenario, synchronization is required because it is a race condition that parallel threads update the value of $w$ and $n$ in a node simultaneously. Enzenberger et. al ignore these race conditions. They accept the possible faulty updates and the inconsistency of parallel computation.      
\item Shared Backup and Selection (SBS): Figure \ref{fig:dr_sbs_1} shows thread $2$ performing \textsc{Backup}($v_{3}$) and thread $3$ performing \textsc{Select}($v_{3}$). In the SBS scenario, synchronization is required because it is a race condition in which a thread reads the value of $w$, and before reading the value of $n$, another thread updates the value of $w$ and $n$. In this case, the first thread reads inconsistent values. Enzenberger et al. ignore these race conditions. They accept the possible faulty updates and the inconsistency of parallel computation.
\end{enumerate}

Algorithm \ref{algo:def_node} shows our new lock-free tree data structure of type \textit{Node} for MCTS. 
Our lock-free tree data structure uses the new multithreading-aware memory model of the C++11 Standard. In order to avoid the  race conditions, the ordering between the memory accesses in the threads has to be enforced \cite{williams2012c++}. In our lock-free implementation, we use the synchronization properties of \textit{atomic} operations to enforce an ordering between the accesses. We have used the atomic variants of the built-in types (i.e., \textit{atomic\_int} and \textit{atomic\_bool}); they are lock-free on most popular platforms. The standard atomic types have different member functions such as \textit{load()}, \textit{store()}, \textit{exchange()}, \textit{fetch\_add()}, and \textit{fetch\_sub()}. The member function \textit{store()} is a store operation, whereas the \textit{load()} is a load operation. The \textit{exchange()} member function replaces the stored value in the atomic variable with a new value and automatically retrieves the original value. We use two memory models for the memory-ordering option for all operations on atomic types: \textit{sequentially consistent} ordering (\textit{memory\_order\_seq\_cst}) and \textit{acquire\_release} 
ordering (\textit{memory\_order\_acquire} and \textit{memory\_order\_release}). The \textit{sequentially consistent} ordering implies that the behavior of a multithreaded program is consistent with a single threaded program. In the  \textit{acquire\_release} ordering model, \textit{load()} is an \textit{acquire} operation, \textit{store()} is a release operation, \textit{exchange()} or \textit{fetch\_add()} or \textit{fetch\_sub()} are either \textit{acquire}, \textit{release} or both (\textit{memory\_order\_acq\_rel}). We have solved all the three above cases of race conditions (SE,SB, and SBS) using these two memory models and the atomic operations. 
\begin{enumerate}
\item  
A node has an $isparent$ flag member. This flag indicates whether the list of children is created or not. The \textit{isparent} flag is initially set to $false$. To change the state of the node to be a parent, we set its $isparent$ to $true$. Before an \textsc{Expand} adds a child to a node, it must create the list of children for the node and set the $isparent$ to $true$. After a node successfully becomes a parent, one of the unexpanded children in this list can be added to the node. It prevents the problem in EE that the list of children is created by two threads at the same time. Thus, the key steps in the \textsc{Expand} operation are as follows: (A, see Algorithm \ref{algo:def_node}) change $v_{6}.isparent$ to $true$ (i.e., no other thread can enter), (B) create the list of children for $v_{6}$, (C) set the value of $v_{6}.untriedmoves$, (D) set the value of $v_{6}.isexpandable$ to $true$ (D1) and load the value of $v_{6}.isexpandable$ (D2), and (E)
$untriedmoves$ is used as a $count$ of the number of items in the list of $children$.
\item In the SB and SBS race conditions, we use atomic types for variables $w$ and $n$. The thread accesses to these variables (reads (F1,F2) and writes (G1,G2)) are  \textit{sequentially consistent}. This memory model preserve the order of operations in all threads. Therefore we have no faulty updates and guarantee consistency of computation. 
\end{enumerate}

\begin{algorithm*}[!t]
\LinesNumbered
\Struct{Node}{
\Type $move$ : int\;
\Type $w$ : atomic\_int\;
\Type $n$ : atomic\_int\;
\Type $isparent := false$ : atomic\_bool\;
\Type $untriedmoves := -1$ : atomic\_int\;
\Type $isexpandable := false$ : atomic\_bool\;
\Type $isfullyexpanded := false$ : atomic\_bool\;
\Type $parent$ : Node*\;
\Type $children$ : Node*[]\;
\Fn{\textit{Init}(moves) : $<$void$>$}{
    int $nomoves$ =  $moves$.size()\;
    \If{!($isparent$.exchange($true$))\Comment*[r]{[see A]}}{ 
        initialize list of $children$ using $moves$\Comment*[r]{[see B]}
        $untriedmoves$.store($nomoves$)\Comment*[r]{[see C]}
        $isexpandable$.store($true$,$memory\_order\_release$)\Comment*[r]{[see D1]}
    }
}
\Fn{\textit{AddChild}() : $<$Node*$>$}{
    int $index$\;
    \uIf{$isexpandable.$load($memory\_order\_acquire$)\Comment*[r]{[see D2]}}{
        \If{($index :=$ $undriedmoves.$fetch\_sub(1)) = 0\Comment*[r]{[see E]}}{
            $isfullyexpanded$.store($true$)\;
        }        
        \uIf{$index$ $<$ 0}{
            \Return current node\;
        }
        \Else{
            \Return $children$[$index$]\;
        }
    }
    \Else{
        \Return current node\;
    }
}
\Fn{\textit{IsFullyExpanded}() : $<$bool$>$}{
    \Return $isfullyexpanded$.load()\;
}
\Fn{\textit{UCT}($C_{p}$) : $<$float$>$}{
    $w^{'}$ := $w$.load($memory\_order\_seq\_cst$)\Comment*[r]{[see F1]}
    $n^{'}$ := $n$.load($memory\_order\_seq\_cst$)\Comment*[r]{[see F2]}
    $n^{"}$ := $parent\rightarrow n$.load($memory\_order\_seq\_cst$)\;
    \Return $\frac{w^{'}}{n^{'}}+C_{p}\sqrt{\frac{\ln(n^{"})}{n^{'}}}$
}
\Fn{\textit{Update}($\Delta$) : $<$void$>$}{
    $w.$fetch\_add($\Delta$,$memory\_order\_seq\_cst$)\Comment*[r]{[see G1]}
    $n.$fetch\_add(1,$memory\_order\_seq\_cst$)\Comment*[r]{[see G2]}
}

}
\caption{Type definition for a lock-free tree data structure.}
\label{algo:def_node}
\end{algorithm*}

\section{Experimental Setup}
\label{sec:setup}
The performance of 3PMCTS is measured by using a High Energy Physics (HEP) expression simplification problem \cite{Kuipers2013}.  
Our setup follows closely \cite{Kuipers2013}. We discuses in \ref{sec:horner} the case study, in \ref{sec:hardware} the hardware, and in \ref{sec:perf_metric} the performance metrics. 

\subsection{Case Study}
\label{sec:horner}

Horner's rule is an algorithm for polynomial computation that reduces the number of multiplications and results in a computationally efficient form. For a polynomial in one variable
\begin{equation}
p(x)=a_{n}x^{n}+a_{n-1}x^{n-1}+...+a_{0},
\end{equation}
the rule simply factors out powers of $x$. Thus, the polynomial can be written in the form
\begin{equation}
p(x)=((a_{n}x+a_{n-1})x+...)x+a_{0}.
\label{hornerrule}
\end{equation}
This representation reduces the number of multiplications to $n$ and has $n$ additions. Therefore, the total evaluation cost of the polynomial is $2n$.

Horner's rule can be generalized for multivariate polynomials. Here, Eq. \ref{hornerrule} applies on a polynomial for each variable, treating the other variables as constants. The order of choosing variables may be different, each order of the variables is called a \textit{Horner scheme}. 

The number of operations can be reduced even more by performing common subexpression elimination (CSE) after transforming a polynomial with Horner's rule. CSE creates new symbols for each subexpression that appears twice or more and replaces them inside the polynomial. Then, the subexpression has to be computed only once.

We are using the HEP($\sigma$) expression with 15 variables to study the results of 3PMCTS. The MCTS is used to find an order of the variables that gives efficient Horner schemes \cite{Kuipers2013}. The root node has $n$ children, with $n$ the number of variables. The children of other nodes represent the remaining unchosen variables in the order. Starting at the root node, a path of nodes (variables) inside the search tree is selected. The incomplete order is completed with the remaining variables added randomly (\textit{RandomSimulation}). This complete order is then used for Horner’s method followed by CSE (\textit{Evaluation}). The number of operations in this optimized expression is counted ($\Delta$).

\subsection{Hardware}
\label{sec:hardware}
Our experiments were performed on a dual socket Intel machine with 2 Intel {\em Xeon\/} E5-2596v2 CPUs running at 2.4 GHz. Each CPU has 12 cores, 24 hyperthreads, and 30 MB L3 cache. Each physical core has 256KB L2 cache. The peak TurboBoost frequency is 3.2 GHz. The machine has 192GB physical memory. We compiled the code using the Intel C++ compiler with a
-$O3$ flag. 

\begin{figure*}[!b]
\begin{tabular}{cc}
\subfloat[CPU]{\includegraphics[width=3.4in]{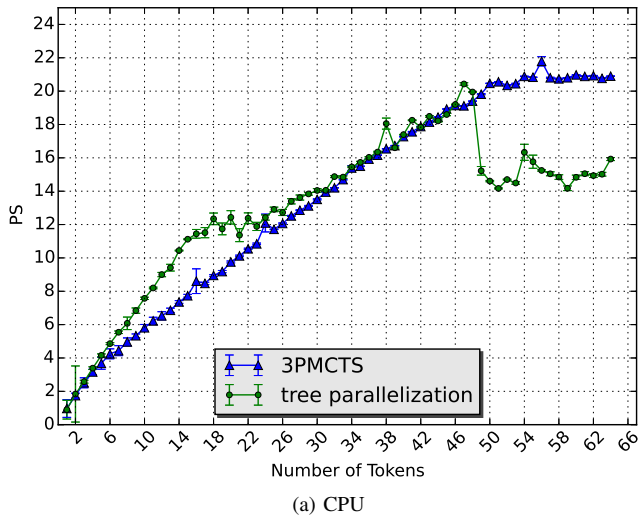}%
\label{fig:playout_speedup}}
&
\subfloat[Phi]{\includegraphics[width=3.35in]{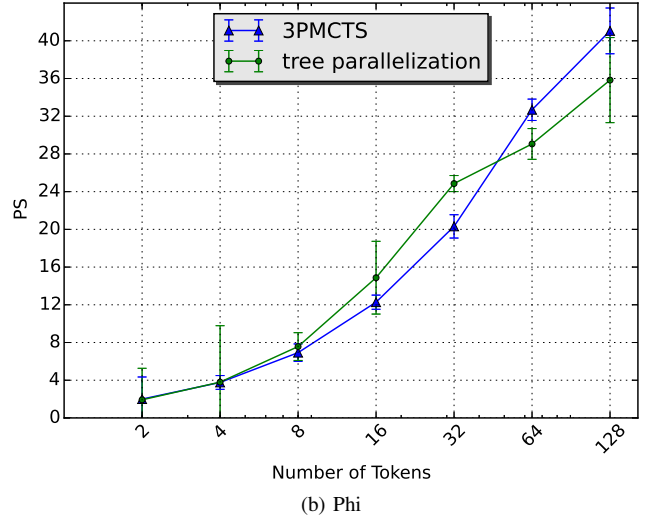}%
\label{fig:playout_speedup_phi}}
\end{tabular}
\caption{ Playout-speedup as function of the number of tokens. Each data point is an average of 10 runs. The constant $C_{p}$ is 0.1. The search budget is 1024 playouts.}
\end{figure*}

\subsection{Performance Metrics}
\label{sec:perf_metric}
The primary goal of parallelization is performance. There are two important metrics related to performance and parallelism for MCTS. 
\begin{enumerate}
\item Playout Speedup (PS): If we have a fixed number of playouts seen as the search budget, then
\begin{equation}
PS = \frac{\text{time in sequential}}{\text{time in parallel}}.
\end{equation}
\item Search Overhead (SO): If we have to find a desired optimal point in the search space, then 
\begin{equation}
SO = \frac{\text{required \# of playouts in parallel}}{\text{required \# of playouts in sequential}} - 1.
\end{equation}

If the parallel MCTS algorithm expands more nodes (i.e., do more playouts) than the equivalent serial algorithm to solve a problem, then there is SO.

\end{enumerate}
In this paper, we use playout-speedup to report the performance.

\section{Experimental Results}
In this section, the performance of 3PMCTS is measured.
\label{sec:results} 
\begin{enumerate}
\item Playout-speedup for CPU:
The graph in Figure \ref{fig:playout_speedup} shows the playout-speedup for both 3PMCTS and tree parallelization, as a function of the number of tokens on CPU. Both 3PMCTS and tree parallelization are doing 1024 playouts. We see a playout-speedup for 3PMCTS close to 22 for 56 tokens. A  playout-speedup close to 21 is observed for tree parallelization for 47 tasks. After 48 tasks the playout-speedup for tree parallelization drops (it is run on a machine with 48 hyperthreads) while the performance of 3PMCTS continues to grow until it becomes stable.

\item Playout-speedup for Phi:
The graph in Figure \ref{fig:playout_speedup_phi} shows the playout-speedup for
both 3PMCTS and tree parallelization, as a function of the number
of tokens on Phi. Both 3PMCTS and tree parallelization are doing 1024 playouts. We see a playout-speedup for 3PMCTS close to 41 for 128 tokens. A  playout-speedup close to 36 is observed for tree parallelization for 128 tasks.
\end{enumerate}
From these results, we may provisionally conclude that the 3PMCTS algorithm shows a playout-speedup as good as tree parallelization, for a well-studied problem. It allows fine-grained managing of the control flow of operations in MCTS in contrast to tree parallelization. 

\section{Related Work}
\label{sec:related}
Below we review related work on MCTS parallelizations.
The two major parallelization methods for MCTS are root parallelization and tree
parallelization~\cite{Chaslot2008}. There exist also less frequently encountered techniques, such as leaf
parallelization~\cite{Chaslot2008} and approaches based on transposition table driven work scheduling (TDS)~\cite{Yoshizoe2011a,Romein99}.
\begin{enumerate}
\item Tree parallelization:
For shared memory machines, tree parallelization is a suitable method. It is used in {\sc Fuego}, a well-known open source Go program \cite{Enzenberger2010b}. In tree parallelization one MCTS tree is shared among several threads that are performing simultaneous searches~\cite{Chaslot2008}. It is shown in~\cite{Chaslot2008} that the playout-speedup of tree parallelization with \textit{virtual loss} does not scale perfectly up to 16 threads. The main challenge is the use of locks to prevent data corruption.

\item Root parallelization:
Chaslot et al.~\cite{Chaslot2008} report that root parallelization shows perfect playout-speedup up to 16 threads.
Soejima et al.~\cite{Soejima2010} analyzed the performance of root parallelization in detail. They revealed that a Go program that uses lock-free tree parallelization with 4 to 8 threads outperformed the same program with root parallelization that utilized 64 distributed CPU cores. This result suggests the superiority of tree parallelization over root parallelization in shared memory machines.
Fern and Lewis ~\cite{fern2011ensemble} thoroughly investigated an Ensemble UCT
approach in which multiple instances of UCT were run independently. 
Their root statistics were combined to yield the final result, showing that Ensembles can significantly improve performance per unit time in a parallel model. This is also shown in \cite{Mirsoleimani2015}.
\end{enumerate}

\section{Conclusion and Future Work}
\label{sec:conclusion}
Monte Carlo Tree Search (MCTS) is a randomized algorithm that is successful in a wide range of optimization problems. The main loop in MCTS consists of individual iterations, suggesting that the algorithm is well suited for parallelization. The existing parallelization methods, e.g., tree parallelization, simply fans out the iterations over available cores.

In this paper, a structured parallel programming approach is used to develop a new parallel algorithm based on the pipeline pattern for MCTS. The idea is to break-up the iterations themselves, splitting them into individual operations, which are then parallelized in a pipeline. Experiments with an application from High Energy Physics show that our implementation of 3PMCTS scales well. Scalability is only one issue,  although it is an important one. The second issue is flexibility of task decomposition in parallelism. These flexibilities allow fine-grained managing of the control flow of operations in MCTS. We consider the flexibility an even more important characteristic of 3PMCTS.

We may conclude the following. Our new method is highly suitable for heterogeneous computing because it is possible that some of the MCTS operations might not be suitable for running on a target processor, whereas others are. Our 3PMCTS algorithm gives us full flexibility for offloading a variety of different operations of MCTS to a target processor. For future work, we will study the implementation of 3PMCTS on a heterogeneous machine.


\pagebreak
\section*{Acknowledgment}
This work is supported in part by the ERC Advanced Grant no. 320651, “HEPGAME.”



\bibliographystyle{IEEEtran}
\bibliography{IEEEabrv,./Bib-icpp2017}

\begin{thebibliography}{10}
\providecommand{\url}[1]{#1}
\csname url@samestyle\endcsname
\providecommand{\newblock}{\relax}
\providecommand{\bibinfo}[2]{#2}
\providecommand{\BIBentrySTDinterwordspacing}{\spaceskip=0pt\relax}
\providecommand{\BIBentryALTinterwordstretchfactor}{4}
\providecommand{\BIBentryALTinterwordspacing}{\spaceskip=\fontdimen2\font plus
\BIBentryALTinterwordstretchfactor\fontdimen3\font minus
  \fontdimen4\font\relax}
\providecommand{\BIBforeignlanguage}[2]{{%
\expandafter\ifx\csname l@#1\endcsname\relax
\typeout{** WARNING: IEEEtran.bst: No hyphenation pattern has been}%
\typeout{** loaded for the language `#1'. Using the pattern for}%
\typeout{** the default language instead.}%
\else
\language=\csname l@#1\endcsname
\fi
#2}}
\providecommand{\BIBdecl}{\relax}
\BIBdecl

\bibitem{Coulom2006}
R.~Coulom, ``{Efficient Selectivity and Backup Operators in Monte-Carlo Tree
  Search},'' in \emph{Proceedings of the 5th International Conference on
  Computers and Games}, ser. CG'06, vol. 4630.\hskip 1em plus 0.5em minus
  0.4em\relax Springer-Verlag, may 2006, pp. 72--83.

\bibitem{Kocsis2006}
L.~Kocsis and C.~Szepesv{\'{a}}ri, ``{Bandit based Monte-Carlo Planning
  Levente},'' in \emph{ECML'06 Proceedings of the 17th European conference on
  Machine Learning}, ser. Lecture Notes in Computer Science,
  J.~F{\"{u}}rnkranz, T.~Scheffer, and M.~Spiliopoulou, Eds., vol. 4212.\hskip
  1em plus 0.5em minus 0.4em\relax Springer Berlin Heidelberg, sep 2006, pp.
  282--293.

\bibitem{Kuipers2013}
J.~Kuipers, A.~Plaat, J.~Vermaseren, and J.~van~den Herik, ``{Improving
  Multivariate Horner Schemes with Monte Carlo Tree Search},'' \emph{Computer
  Physics Communications}, vol. 184, no.~11, pp. 2391--2395, nov 2013.

\bibitem{goodfellow2016deep}
\BIBentryALTinterwordspacing
I.~Goodfellow, Y.~Bengio, and A.~Courville, \emph{Deep Learning}, ser. Adaptive
  Computation and Machine Learning Series.\hskip 1em plus 0.5em minus
  0.4em\relax MIT Press, 2016. [Online]. Available:
  \url{https://books.google.nl/books?id=Np9SDQAAQBAJ}
\BIBentrySTDinterwordspacing

\bibitem{Chaslot2008}
G.~Chaslot, M.~Winands, and J.~van~den Herik, ``{Parallel Monte-Carlo Tree
  Search},'' in \emph{the 6th Internatioal Conference on Computers and Games},
  vol. 5131.\hskip 1em plus 0.5em minus 0.4em\relax Springer Berlin Heidelberg,
  2008, pp. 60--71.

\bibitem{Yoshizoe2011a}
K.~Yoshizoe, A.~Kishimoto, T.~Kaneko, H.~Yoshimoto, and Y.~Ishikawa,
  ``\BIBforeignlanguage{en}{{Scalable Distributed Monte-Carlo Tree Search}},''
  in \emph{\BIBforeignlanguage{en}{Fourth Annual Symposium on Combinatorial
  Search}}, may 2011, pp. 180--187.

\bibitem{fern2011ensemble}
A.~Fern and P.~Lewis, ``{Ensemble Monte-Carlo Planning: An Empirical Study.}''
  in \emph{ICAPS}, 2011, pp. 58--65.

\bibitem{Schaefers2014}
L.~Schaefers and M.~Platzner, ``{Distributed Monte-Carlo Tree Search : A Novel
  Technique and its Application to Computer Go},'' \emph{IEEE Transactions on
  Computational Intelligence and AI in Games}, vol.~6, no.~3, pp. 1--15, 2014.

\bibitem{Mirsoleimani2015a}
S.~A. Mirsoleimani, A.~Plaat, J.~van~den Herik, and J.~Vermaseren, ``{Parallel
  Monte Carlo Tree Search from Multi-core to Many-core Processors},'' in
  \emph{ISPA 2015 : The 13th IEEE International Symposium on Parallel and
  Distributed Processing with Applications (ISPA)}, Helsinki, 2015, pp. 77--83.

\bibitem{Mirsoleimani2015}
------, ``{Scaling Monte Carlo Tree Search on Intel Xeon Phi},'' in
  \emph{Parallel and Distributed Systems (ICPADS), 2015 20th IEEE International
  Conference on}, 2015, pp. 666--673.

\bibitem{tbb2015}
\BIBentryALTinterwordspacing
``{Intel threading building blocks TBB}.'' [Online]. Available:
  \url{https://www.threadingbuildingblocks.org}
\BIBentrySTDinterwordspacing

\bibitem{Browne2012}
C.~B. Browne, E.~Powley, D.~Whitehouse, S.~M. Lucas, P.~I. Cowling,
  P.~Rohlfshagen, S.~Tavener, D.~Perez, S.~Samothrakis, and S.~Colton, ``{A
  Survey of Monte Carlo Tree Search Methods},'' \emph{Computational
  Intelligence and AI in Games, IEEE Transactions on}, vol.~4, no.~1, pp.
  1--43, 2012.

\bibitem{Enzenberger2010a}
M.~Enzenberger and M.~M{\"{u}}ller, ``{A lock-free multithreaded Monte-Carlo
  tree search algorithm},'' \emph{Advances in Computer Games}, vol. 6048, pp.
  14--20, 2010.

\bibitem{tbb-reinders2007-book}
J.~Reinders, \emph{{Intel threading building blocks: outfitting C++ for
  multi-core processor parallelism}}.\hskip 1em plus 0.5em minus 0.4em\relax "
  O'Reilly Media, Inc.", 2007.

\bibitem{cilk1998}
C.~E. Leiserson and A.~Plaat, ``{Programming Parallel Applications in Cilk},''
  \emph{SINEWS: SIAM News}, vol.~31, no.~4, pp. 6--7, 1998.

\bibitem{McCool2012}
M.~McCool, J.~Reinders, and A.~Robison, \emph{{Structured Parallel Programming:
  Patterns for Efficient Computation}}.\hskip 1em plus 0.5em minus 0.4em\relax
  Elsevier, 2012.

\bibitem{Enzenberger2010b}
M.~Enzenberger, M.~Muller, B.~Arneson, and R.~Segal,
  ``\BIBforeignlanguage{English}{{Fuego—An Open-Source Framework for Board
  Games and Go Engine Based on Monte Carlo Tree Search}},''
  \emph{\BIBforeignlanguage{English}{IEEE Transactions on Computational
  Intelligence and AI in Games}}, vol.~2, no.~4, pp. 259--270, dec 2010.

\bibitem{williams2012c++}
\BIBentryALTinterwordspacing
A.~Williams, \emph{C++ Concurrency in Action: Practical Multithreading}, ser.
  Manning Pubs Co Series.\hskip 1em plus 0.5em minus 0.4em\relax Manning, 2012.
  [Online]. Available: \url{https://books.google.nl/books?id=EttPPgAACAAJ}
\BIBentrySTDinterwordspacing

\bibitem{Romein99}
J.~Romein, A.~Plaat, H.~E. Bal, and J.~Schaeffer, ``{Transposition Table Driven
  Work Scheduling in Distributed Search},'' in \emph{In 16th National
  Conference on Artificial Intelligence (AAAI'99)}, 1999, pp. 725--731.

\bibitem{Soejima2010}
Y.~Soejima, A.~Kishimoto, and O.~Watanabe, ``{Evaluating Root Parallelization
  in Go},'' \emph{IEEE Transactions on Computational Intelligence and AI in
  Games}, vol.~2, no.~4, pp. 278--287, dec 2010.

\end{thebibliography}
%
%
%

\end{document}